\documentclass{article}
\usepackage{arxiv}
\pdfoutput=1
\usepackage[numbers]{natbib}

\usepackage[utf8]{inputenc} 
\usepackage[T1]{fontenc}    
\usepackage{hyperref}       
\usepackage{url}            
\usepackage{booktabs}       
\usepackage{amsfonts}       
\usepackage{nicefrac}       
\usepackage{microtype}      
\usepackage{graphicx}
\title{To Tune or Not To Tune? How About the Best of Both Worlds?}

\author{
  Ran Wang \\
  AI Lab, Percent Group \\
  Beijing, Chaoyang District \\
  \texttt{lucien.wang@percent.cn} \\
  \And
  Haibo Su \\
  AI Lab, Percent Group \\
  Beijing, Chaoyang District \\
  \texttt{haibo.su@percent.cn} \\
  \And
  Chunye Wang \\
  AI Lab, Percent Group \\
  Beijing, Chaoyang District \\
  AI Lab, Percent Group \\
  \texttt{chunye.wang@percent.cn} \\
  \And 
  Kailin Ji \\
  AI Lab, Percent Group \\
  Beijing, Chaoyang District \\
  \texttt{kailin.ji@percent.cn} \\
  \And
  Jupeng Ding \\
  AI Lab, Percent Group \\
  Beijing, Chaoyang District \\
  \texttt{jupeng.ding@percent.cn} 
}

\begin{document}

\maketitle

\begin{abstract}
The introduction of pre-trained language models has revolutionized natural language research communities. However, researchers still know relatively little regarding their theoretical and empirical properties. In this regard, Peters et al.\cite{peters2019tune} performs several experiments which demonstrate that it is better to adapt BERT with a light-weight task-specific head, rather than building a complex one on top of the pre-trained language model, and freeze the parameters in the said language model. However, there is another option to adopt. In this paper, we propose a new adaptation method which we first train the task model with the BERT parameters frozen and then fine-tune the entire model together. Our experimental results show that our model adaptation method can achieve 4.7\% accuracy improvement in semantic similarity task, 0.99\% accuracy improvement in sequence labeling task and 0.72\% accuracy improvement in the text classification task.
\end{abstract}

\section{Introduction}

The introduction of pre-trained language models, such as BERT \cite{devlin2018bert} and Open-GPT \cite{radford2019language}, among many others, has brought tremendous progress to the NLP research and industrial communities. The contribution of these models can be categorized into two aspects. First, pre-trained language models allow modelers to achieve reasonable accuracy without the need an excessive amount of manually labeled data. This strategy is in contrast with the classical deep learning methods, which requires a multitude more data to reach comparable results. Second, for many NLP tasks, including but not limited to,  SQuAD \cite{rajpurkar2016squad}, CoQA \cite{reddy2018coqa}, named entity recognition \cite{sang2003introduction}, Glue \cite{wang2018glue}, machine translation \cite{jean2015montreal}, pre-trained model allows the creation of new state-of-art, given a reasonable amount of labelled data. 

In the post pre-trained language model era, to pursue new state-of-art, two directions can be followed. The first method, is to improve the pre-training process, such as in the work of ERNIE \cite{sun2019ernie}, GPT2.0 \cite{radford2019language} and MT-DNN \cite{liu2019multi}. The second method is to stand on the shoulder of the pre-trained language models. Among the many possibilities, one of them is to build new neural network structures on top of pre-trained language models. 

In principles, there are three ways to train the networks with stacked neural networks on top of pre-trained language models, as shown in Table \ref{pretrain}. In Peters et al . \cite{peters2019tune}, the authors compare the possibility of option \textit{stack-only} and \textit{finetune-only}, and conclude that option \textit{finetune-only} is better than option \textit{stack-only}. More specifically, Peter et al. \cite{peters2019tune} argue that it is better to add a task-specific head on top of BERT than to freeze the weights of BERT and add more complex network structures. However, Peters et al. \cite{peters2019tune} did not compare option \textit{stack-and-finetune} and \textit{finetune-only}. On the other hand, before pre-trained deep language models became popular, researchers often use a strategy analog to option \textit{stack-and-finetune}. That is, modelers first train the model until convergence, and then fine-tune the word embeddings with a few epochs. If pre-trained language models can be understood as at least partially resemblance of word embeddings, then it will be imprudent not to consider the possibility of option \textit{stack-and-finetune}. 

In this study, we aim to compare the strategy \textit{stack-and-finetune} and strategy \textit{finetune-only}. More specifically, we perform three NLP tasks, sequence labeling, text classification, and question similarity. In the first tasks, we demonstrate that even without modifying the network structures, building networks on top of pre-trained language models might improve accuracy. In the second tasks, we show that by ensembling different neural networks, one can even improve the accuracy of fine-tuning only methods even further. Finally, in the last task, we demonstrate that if one can tailor-made a neural network that specifically fit the characteristics of the pre-trained language models, one can improve the accuracy even further.  All the results indicate the strategy \textit{stack-and-finetune} is superior to strategy \textit{finetune-only}. This leads us to conclude that, at least, by overlooking the possibility strategy \textit{stack-and-finetune}  is imprudent.

\begin{table}[]
\centering
\begin{tabular}{|l|l|l|}
\hline
                                                                                           & \textit{\textbf{\begin{tabular}[c]{@{}l@{}}Fine-tune \\ pretrained model\end{tabular}}} & \textit{\textbf{\begin{tabular}[c]{@{}l@{}}Do not fine-tune \\ pretrained model\end{tabular}}} \\ \hline
\textit{\textbf{\begin{tabular}[c]{@{}l@{}}Train the model \\ on top\end{tabular}}}        & Stack-and-finetune                                                                                    & Stack-only                                                                                              \\ \hline
\textit{\textbf{\begin{tabular}[c]{@{}l@{}}Do not train \\the model on top\end{tabular}}} &   Finetune-only                                                                                    & Not reasonable                                                                                 \\ \hline
\end{tabular}
\caption{Methods to Stack Neural Networks on Top of Pre-trained Language Models}
\label{pretrain}
\end{table}

The contribution of this paper is two-fold. First, we propose a new strategy to improve the fine-tune-only strategy proposed by Peter et al. \cite{peters2019tune}, this allows us to achieve better results, at least on the selected tasks. More importantly, the results of this study demonstrate the importance of neural networks design, even in the presence of all-powerful pre-trained language models.  Second, during the experiment, we have found that although simply using the proposed training strategy can result in higher accuracies compared to that of Peter et al. \cite{peters2019tune}, it is still a challenging task to find the appropriate methods to design and to utilize pre-trained networks. In this regard, we find that pre-trained models differ significantly from word embeddings in terms of their training strategies. Especially, since word embeddings can be viewed as shallow transfer learning, while pre-trained model should be viewed as deep transfer learning, one must try to combat over-fitting problems with more care due to the enormous number of parameters presented in the pre-trained models.  Besides, we also find that in order to achieve the maximal performance in the post-pre-trained language model era, one must design, either manually or via Auto ML,  networks that best fit the structure, especially the depth of the pre-trained language models. 

The rest of the paper is organized as follows. First, we review the relevant literature on pre-trained deep neural networks, the argument in Peter et al. \cite{peters2019tune} as well as fine-tuning strategies with word embeddings. Second, we present three experiments and showed the superiority of strategy \textit{stack-and-finetune} compared to strategy \textit{finetune-only}. Finally, we conclude with some remarks and future research possibilities. 

\section{Related Studies}
Before the introduction of deep neural networks, researchers in the field of NLP have been using pre-trained models. Among all of them, one of the most famous is the word embeddings, which maps each word into a continuous vector, instead of one-hot encodings \cite{mikolov2013efficient}. By doing so, not only are we able to reduce the dimensionality of the input features, which helps to avoid over-fitting, but also capture, at least partially, the internal meaning of each word. 

However, since each word is only endowed with a fixed numerical vector in the methodology of  word embeddings, word embeddings are unable to capture the contextual meaning in the text. For example, consider the word ''bank'' sentences ``I am walking on the bank of the river.'' with ``I am going to rob the bank''. It is obvious that the word ``bank'' represents completely different meaning, which the word embeddings techniques fail to capture. 

\begin{figure}[!htpb]
\label{bert}
\centering
\includegraphics[scale=1.2]{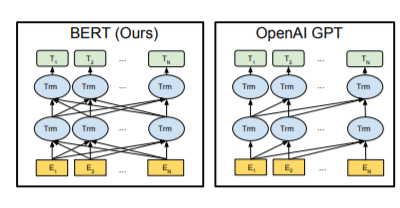}
\caption{The Difference Between BERT and Open-GPT, extracted from Devlin et al. \cite{devlin2018bert}, Figure 1}
\end{figure}

The aforementioned deficiencies prompt researchers to propose deep neural networks that are able to be trained in an unsupervised fashion while being able to capture the contextual meaning of the 
words presented in the texts. Some early attempts include pre-trained models includes, CoVe \cite{mccann2017learned}, CVT \cite{he2016multi, clark2018semi}, ELMo \cite{peters2018deep} and ULMFiT \cite{howard2018universal}. However, the most successful ones are BERT \cite{devlin2018bert} and Open-GPT \cite{radford2019language}. Unlike standard NLP deep learning model, BERT and Open-GPT are built on top of transformer \cite{vaswani2017attention} structures, instead of LSTM \cite{hochreiter1997long} or GRU \cite{chung2015gated}. The difference between BERT and Open-GPT is that BERT uses bi-directional self-attentions while Open-GPT uses only unidirectional ones, as shown in Figure \ref{bert}. The transformer structures differ from the LSTM's in the two important aspects. First, it allows for stacking of multiple layers with residual connections and batch normalizations, which allows for free gradient flow. Second, the core computational unit is matrix multiplications, which allows researchers to utilize the full computational potential of TPU \cite{sato2017depth}.
After training on a large corpus, both BERT and Open-GPT are able to renew the SOTA of many important natural language tasks, such as such as SQuAD \cite{rajpurkar2016squad}, CoQA \cite{reddy2018coqa}, named entity recognition \cite{sang2003introduction}, Glue \cite{wang2018glue}, machine translation \cite{jean2015montreal}. 

In the presence of the success of pre-trained language models, especially BERT \cite{devlin2018bert},  it is natural to ask how to best utilize the pre-trained language models to achieve new state-of-the-art results. In this line of work, Liu et al. \cite{liu2019linguistic} investigated the linguistic knowledge and transferability of contextual representations by comparing BERT \cite{devlin2018bert} with ELMo \cite{peters2018deep}, and concluded that while the higher levels of LSTM's are more task-specific, this trend does not exhibit in transformer based models. Stickland and Murray \cite{stickland2019bert} invented projected attention layer for multi-task learning using BERT, which results in an improvement in various state-of-the-art results compared to the original work of Devlin et al. \cite{devlin2018bert}. Xu et al. \cite{xu2019bert} propose a ``post-training'' algorithms, which does not directly fine-tune BERT, but rather first ``post-train'' BERT on the task related corpus using the masked language prediction task next sentence prediction task, which helps to reduce the bias in the training corpus. Finally, Sun et al. \cite{sun2019fine} added additional fine-tuning tasks based on multi-task training, which further improves the prediction power of BERT in the tasks of text classification. 

In this aspect, however, there is a simple yet crucial question that needs to be addressed. That is, whether it is possible to top BERT with the commonly used or task specific layers, and if this is possible, how to best utilize the pre-trained language models in this situation. In this regards, Peters et al. \cite{peters2019tune} investigated how to best adapt the pre-trained model to a specific task, and focused on two different adaptation method,feature extraction and directly fine-tuning the pre-trained model, which corresponding to the strategy \textit{finetune-only} and the strategy \textit{stack-only} in Table \ref{pretrain}. On this regard, Peters et al. \cite{peters2019tune} performs five experiments, including: (1) named entity recognition \cite{sang2003introduction}; (2) sentiment analysis \cite{socher2013recursive}; (3) natural language inference \cite{williams2017broad}; (4) paraphrase detection \cite{dolan2005automatically}; (5) semantic textual similarity \cite{cer2017semeval}. By the results of these tasks, Peters et al. \cite{peters2019tune} concludes that adding a light task-specific head and performing fine-tuning on BERT is better than building a complex network on top without BERT fine-tuning.

\section{Methodology}
Under our strategy \textit{stack-and-finetune}, the model training process is divided into two phases, which are described in detail below. In the first phase, the parameters of the pre-training model are fixed, and only the upper-level models added for a specific task is learned. In the second phase, we fine-tune the upper-level models together with the pre-trained language models. We choose this strategy for the following reasons.  Pre-training models have been used to obtain more effective word representations through the study of a large number of corpora.  In the paradigm proposed in the original work by Devlin et al. \cite{devlin2018bert}, the author directly trained BERT along with with a light-weighted task-specific head. In our case though, we top BERT with a more complex network structure, using Kaiming initialization \cite{he2015delving}. If one would fine-tune directly the top models along with the weights in BERT, one is faced with the following dilemma: on the one hand, if the learning rate is too large, it is likely to disturb the structure innate to the pre-trained language models; on the other hand, if the learning rate is too small, since we top BERT with relatively complex models, the convergence of the top models might be impeded. Therefore, in the first phase we fix the weights in the pre-training language models, and only train the model on top of it. 

Another aspect that is worth commenting in the first phase is that it is most beneficial that one does not train the top model until it reaches the highest accuracy on the training or validation data sets, but rather only up to a point where the prediction accuracy of the training and validation data sets do not differ much. This is intuitively reasonable for the following reasons. Unlike word embeddings, the pre-trained language models possess a large number of parameters compared to the task-specific models we build on top them. Therefore, if one were to train the top models until they reach the highest prediction accuracy in the training or validation data sets, it would likely cause the models to over-fit. Therefore, in our experiment, we found that this leads to the highest performance increase in the fine-tuning stage.

\section{Experiments}
\subsection{Overview}
We perform three different experiments to test our hypotheses. First, we perform a named entity recognition tasks, by adding a bi-LSTM on top of the BERT model. In this experiment, we hope to test whether, without any modification to the commonly used network structure, our proposed training strategy will improve the overall accuracy. Second, we perform a text classification experiments, in this experiments, we trained three models, and perform a model ensemble. We hope to show that even the added network has not contributed to significantly in improving the accuracy, it does provide opportunities for model ensembles. Finally, we perform the textual similarity tests, in which we show that if one can tailor make a network that specifically fit the characteristics of the pre-trained languages, more significant improvement can be expected.

Under the strategy \textit{finetune-only}, we use only single BERT.In order to adapt to different tasks, we will add a fully connected layer upon BERT. In the sequence labeling task, the BERT word embedding of each word passes through two fully connected layers, and the prediction probability of named entity can be obtained. In the next two verification tasks, we use ``[CLS]'' for prediction and add two fully connected layers subsequently. Under our strategy \textit{stack-and-finetune}, we set different learning rates for the two phases. We tried to set the learning rate of the first stage to $1e^{-1}$,$1e^{-2}$,$5e^{-3}$,$1e^{-3}$ and $5e^{-4}$, and set it to a smaller number in the latter stage, such as $1e^{-3}$,$1e^{-4}$ ,$5e^{-5}$ and $1e^{-5}$. After our experiments, we found that it gets better results while the learning rate is set to 0.001 in the stage of training only the upper model and set to $5e^{-5}$  in the later stage. Since BERT-Adam\cite{devlin2018bert} has excellent performance, in our experiments, we use it as an optimizer with $\beta_{1} = 0.9,\beta_{2} = 0.999$,$L_2$-weight decay of $0.01$.We apply a dropout trick on all layers and set the dropout probability as 0.1.

\subsection{Experiment A: Sequence Labeling}
In the sequence labeling task,we explore sub-task named entity recognition using CoNLL03 dataset  \cite{sang2003introduction}, which is a public available used in many studies to test the accuracy of their proposed methods \cite{pennington2014glove, chiu2016named, lample2016neural, ma2016end, devlin2018bert}. For strategy \textit{finetune-only} and strategy \textit{stack-and-finetune}, we implemented two models: one with BERT and the other with BERT adding a Bi-LSTM on top. Eval measure is accuracy and F1 score.
\begin{table}[!h]
\label{ner}
\centering
    {\begin{tabular}{ccc}
    \hline
    Model& Accuracy(\%) & F1(\%)\\
    \hline
    BERT& 98.75 & 85.62\\
    BERT + Bi-LSTM & \textbf{99.74} & \textbf{92.51}\\
    \hline
    \end{tabular}}
    \caption{Results for named entity recognition}
\end{table}

As is shown in Table 2, even without modifying the networks to specifically adapt to the pre-trained model, our training strategy still brought improvement towards overall accuracy of 0.99\% for the accuracy and 0.068 on the F1 score, proving the success of our proposed methods.

\subsection{Experiment B: Text Classification}

In the task of text categorization, we used Yahoo Answer Classification Dataset. The Dataset is consists of 10 classes, but due to the huge amount of the dataset, we just select two class of them. As for the upper model,we choose DenseNet\cite{le2018convolutional} and HighwayLSTM \cite{zhang2016highway}.

The DenseNet structure contains four independent blocks and each block has four CNNs connected by residual. We initialize word embedding in the word representation layer with BERT. We initialize each character as a 768-dimension vector. In the experiment of training DenseNet,we concat the output vector of DenseNet with [CLS] for prediction. 

\begin{table}[!h]
\centering

    {\begin{tabular}{cccc}
    \hline
     Model& Accuracy(\%) & F1(\%)\\
    \hline
    BERT& 90.05 &89.97\\
    BERT + HighwayLSTM &89.85 &89.85 \\
    BERT + DenseNet& 90.31 &90.25 \\
    Ensembled Model & \textbf{90.77} & \textbf{90.43} \\ 
    \hline
    \end{tabular}}
\caption{Results for text classification}   
\end{table}

We find the ensembled model enjoys a 0.72\% improvements compared to the fine-tune only model and 0.005 improvement for the F1 score. 

\subsection{Experiment C: Semantic Similarity Tasks}

We use ``Quora-Question-Pair'' dataset  \href{https://data.quora.com/First-Quora-Dataset-Release-Question-Pairs}{1}. This is a commonly used dataset containing 400k question pairs, annotated manually to be semantically equivalent or not. Due to its high quality, it is a standard dataset to test the success of various semantic similarity tasks. Various models which are tested on this data set are proposed, including but not limited to \cite{mueller2016siamese, wang2016compare, chen2016enhanced, wang2017bilateral}.

Apart from the BERT fine-tuning only model and BERT+ BIMPM model, we also devise two new network structures by modifying the BIMPM model. In the first model is to remove the first bi-LSTM of BIMPM, which is the input layer for the matching layer in BIMPM. In the second model, we combine the matching layer of BIMPM and with a transformer\cite{vaswani2017attention}, a model we call \textit{Sim-Transformer} by replacing the output layer of the matching layer, originally a bi-LSTM model, with a transformer model. From the experimental results shown in Table 4, we can see that due to the strong expressive ability of the BERT, there is almost no difference in the experimental results of removing the first bi-LSTM and BIMPM. In addition, we also find that Sim-Transformer's performance without fine-tuning is nearly four percentage points lower than BIMPM, but it out-performs BIMPM after fine-tuning. In general, the results show that BERT + Sim-Transformer out-performs BERT-only model by 4.7\%, thus confirming our hypotheses again. 
\begin{table}[!h]
\centering
    {\begin{tabular}{ccc}
    \hline
    Model& Accuracy(BERT fixed) & Accuracy(Fine-tune BERT)\\
    \hline
    BERT& -- & 0.8445\\
    BERT + BIMPM \cite{wang2017bilateral}& \textbf{0.8739} & 0.8899 \\
    BERT + BIMPM(First bi-LSTM removed) & 0.869 & 0.8855 \\
    BERT + Sim-Transformer & 0.8341& \textbf{0.8915} \\
    \hline
    \end{tabular}}
    \label{quora}
    \caption{Results for semantic similarity task}
\end{table}

\section{Discussions and Conclusions}
In summary, we find that in all the three tasks, our proposed method out-performs the methods of simply tuning pre-trained language models, as is proposed in \cite{peters2019tune}. However, we would like to caution the readers in two aspects when reading the conclusion of this study. First, this study does not argue that our proposed methods are \textit{always} superior to fine-tuning only methods. For example, all the experiments in our study are based on data sets of relatively large size. In the other spectrum, if one is only given a limited data set, then building complex networks upon pre-trained language models might lead to disastrous over-fitting. If this is the case, then it is possible that deep domain adaptation \cite{wang2018deep} might be a better choice if one desires to stack neural networks on top of pre-trained language models. However, most domain adaptation applications belong to the field of computer vision, therefore, a call for domain adaptations research in the NLP fields. 

During the experimentation, we also discover some tricks to obtain higher quality networks. The first is that due to the enormous number of parameters presented in the pre-trained language models, to achieve generalizable results on the test data sets, it is vital to combat over-fitting. In classical embedding + training networks, the general training method is to fix the word-embeddings, then train the top model until it converges, and finally fine-tuning the word-embeddings for a few epochs. This training strategy does not work when we replace pre-trained language models with word-embeddings. In our experiment, we first fix the pre-trained language models, and then we train the top neural networks only for a few epochs, until it reaches a reasonable accuracy, while closely monitoring the discrepancy between training accuracy and testing accuracy. After that, we fine-tune the pre-trained language model as well as our models on top together. This allows us to achieve better results on the experimentation. However, it is not yet clear to us when to stop the training of top neural networks. This poses an even more essential question for Auto ML researchers in the following sense. In the classical computer vision based Auto ML approaches, since one seldom build networks on already trained models, there is no particular need to auxiliary measure for over-fittings. While if Auto ML is to be performed on NLP tasks successfully, it might be essential that the gap between training accuracy and test accuracy to be incorporated when one evaluates the model. 

Finally, it is not yet clear what is the most proper way to build networks that tops the pre-trained language models. However, there are several principles that we can follow when designing such networks. First, such networks must be able to ensure the gradient flow from the top of the model to the bottom. This is essential due to the depth of the pre-trained language model. Second, this also means, one does not need explicitly to build extremely complex networks on top of pre-trained language models unless it complements the mechanisms of self-attention. Finally, a challenge remains as to how to use the \textit{depth} of pre-trained language models. The process of our experiment shows that utilizing deeper layers might be a fruitful way to achieve better accuracy.

\bibliographystyle{unsrt}  
\bibliography{tune}

\begin{thebibliography}{10}

\bibitem{peters2019tune}
Matthew Peters, Sebastian Ruder, and Noah~A Smith.
\newblock To tune or not to tune? adapting pretrained representations to
  diverse tasks.
\newblock {\em arXiv preprint arXiv:1903.05987}, 2019.

\bibitem{devlin2018bert}
Jacob Devlin, Ming-Wei Chang, Kenton Lee, and Kristina Toutanova.
\newblock Bert: Pre-training of deep bidirectional transformers for language
  understanding.
\newblock {\em arXiv preprint arXiv:1810.04805}, 2018.

\bibitem{radford2019language}
Alec Radford, Jeffrey Wu, Rewon Child, David Luan, Dario Amodei, and Ilya
  Sutskever.
\newblock Language models are unsupervised multitask learners.
\newblock {\em OpenAI Blog}, 1:8, 2019.

\bibitem{rajpurkar2016squad}
Pranav Rajpurkar, Jian Zhang, Konstantin Lopyrev, and Percy Liang.
\newblock Squad: 100,000+ questions for machine comprehension of text.
\newblock {\em arXiv preprint arXiv:1606.05250}, 2016.

\bibitem{reddy2018coqa}
Siva Reddy, Danqi Chen, and Christopher~D Manning.
\newblock Coqa: A conversational question answering challenge.
\newblock {\em arXiv preprint arXiv:1808.07042}, 2018.

\bibitem{sang2003introduction}
Erik~F Sang and Fien De~Meulder.
\newblock Introduction to the conll-2003 shared task: Language-independent
  named entity recognition.
\newblock {\em arXiv preprint cs/0306050}, 2003.

\bibitem{wang2018glue}
Alex Wang, Amapreet Singh, Julian Michael, Felix Hill, Omer Levy, and Samuel~R
  Bowman.
\newblock Glue: A multi-task benchmark and analysis platform for natural
  language understanding.
\newblock {\em arXiv preprint arXiv:1804.07461}, 2018.

\bibitem{jean2015montreal}
S{\'e}bastien Jean, Orhan Firat, Kyunghyun Cho, Roland Memisevic, and Yoshua
  Bengio.
\newblock Montreal neural machine translation systems for wmt’15.
\newblock In {\em Proceedings of the Tenth Workshop on Statistical Machine
  Translation}, pages 134--140, 2015.

\bibitem{sun2019ernie}
Yu~Sun, Shuohuan Wang, Yukun Li, Shikun Feng, Xuyi Chen, Han Zhang, Xin Tian,
  Danxiang Zhu, Hao Tian, and Hua Wu.
\newblock Ernie: Enhanced representation through knowledge integration.
\newblock {\em arXiv preprint arXiv:1904.09223}, 2019.

\bibitem{liu2019multi}
Xiaodong Liu, Pengcheng He, Weizhu Chen, and Jianfeng Gao.
\newblock Multi-task deep neural networks for natural language understanding.
\newblock {\em arXiv preprint arXiv:1901.11504}, 2019.

\bibitem{mikolov2013efficient}
Tomas Mikolov, Kai Chen, Gregory~S Corrado, and Jeffrey Dean.
\newblock Efficient estimation of word representations in vector space.
\newblock {\em international conference on learning representations}, 2013.

\bibitem{mccann2017learned}
Bryan McCann, James Bradbury, Caiming Xiong, and Richard Socher.
\newblock Learned in translation: Contextualized word vectors.
\newblock In {\em Advances in Neural Information Processing Systems}, pages
  6294--6305, 2017.

\bibitem{he2016multi}
Wanjia He, Weiran Wang, and Karen Livescu.
\newblock Multi-view recurrent neural acoustic word embeddings.
\newblock {\em arXiv preprint arXiv:1611.04496}, 2016.

\bibitem{clark2018semi}
Kevin Clark, Minh-Thang Luong, Christopher~D Manning, and Quoc~V Le.
\newblock Semi-supervised sequence modeling with cross-view training.
\newblock {\em arXiv preprint arXiv:1809.08370}, 2018.

\bibitem{peters2018deep}
Matthew~E Peters, Mark Neumann, Mohit Iyyer, Matt Gardner, Christopher~G Clark,
  Kenton Lee, and Luke~S Zettlemoyer.
\newblock Deep contextualized word representations.
\newblock {\em north american chapter of the association for computational
  linguistics}, 1:2227--2237, 2018.

\bibitem{howard2018universal}
Jeremy Howard and Sebastian Ruder.
\newblock Universal language model fine-tuning for text classification.
\newblock {\em meeting of the association for computational linguistics},
  1:328--339, 2018.

\bibitem{vaswani2017attention}
Ashish Vaswani, Noam Shazeer, Niki Parmar, Jakob Uszkoreit, Llion Jones,
  Aidan~N Gomez, {\L}ukasz Kaiser, and Illia Polosukhin.
\newblock Attention is all you need.
\newblock In {\em Advances in neural information processing systems}, pages
  5998--6008, 2017.

\bibitem{hochreiter1997long}
Sepp Hochreiter and J{\"u}rgen Schmidhuber.
\newblock Long short-term memory.
\newblock {\em Neural computation}, 9(8):1735--1780, 1997.

\bibitem{chung2015gated}
Junyoung Chung, Caglar Gulcehre, Kyunghyun Cho, and Yoshua Bengio.
\newblock Gated feedback recurrent neural networks.
\newblock In {\em International Conference on Machine Learning}, pages
  2067--2075, 2015.

\bibitem{sato2017depth}
Kaz Sato, Cliff Young, and David Patterson.
\newblock An in-depth look at google’s first tensor processing unit (tpu).
\newblock {\em Google Cloud Big Data and Machine Learning Blog}, 12, 2017.

\bibitem{liu2019linguistic}
Nelson~F Liu, Matt Gardner, Yonatan Belinkov, Matthew Peters, and Noah~A Smith.
\newblock Linguistic knowledge and transferability of contextual
  representations.
\newblock {\em arXiv preprint arXiv:1903.08855}, 2019.

\bibitem{stickland2019bert}
Asa~Cooper Stickland and Iain Murray.
\newblock Bert and pals: Projected attention layers for efficient adaptation in
  multi-task learning.
\newblock {\em arXiv preprint arXiv:1902.02671}, 2019.

\bibitem{xu2019bert}
Hu~Xu, Bing Liu, Lei Shu, and Philip~S Yu.
\newblock Bert post-training for review reading comprehension and aspect-based
  sentiment analysis.
\newblock {\em arXiv preprint arXiv:1904.02232}, 2019.

\bibitem{sun2019fine}
Chi Sun, Xipeng Qiu, Yige Xu, and Xuanjing Huang.
\newblock How to fine-tune bert for text classification?
\newblock {\em arXiv preprint arXiv:1905.05583}, 2019.

\bibitem{socher2013recursive}
Richard Socher, Alex Perelygin, Jean Wu, Jason Chuang, Christopher~D Manning,
  Andrew Ng, and Christopher Potts.
\newblock Recursive deep models for semantic compositionality over a sentiment
  treebank.
\newblock In {\em Proceedings of the 2013 conference on empirical methods in
  natural language processing}, pages 1631--1642, 2013.

\bibitem{williams2017broad}
Adina Williams, Nikita Nangia, and Samuel~R Bowman.
\newblock A broad-coverage challenge corpus for sentence understanding through
  inference.
\newblock {\em arXiv preprint arXiv:1704.05426}, 2017.

\bibitem{dolan2005automatically}
William~B Dolan and Chris Brockett.
\newblock Automatically constructing a corpus of sentential paraphrases.
\newblock In {\em Proceedings of the Third International Workshop on
  Paraphrasing (IWP2005)}, 2005.

\bibitem{cer2017semeval}
Daniel Cer, Mona Diab, Eneko Agirre, Inigo Lopez-Gazpio, and Lucia Specia.
\newblock Semeval-2017 task 1: Semantic textual similarity-multilingual and
  cross-lingual focused evaluation.
\newblock {\em arXiv preprint arXiv:1708.00055}, 2017.

\bibitem{he2015delving}
Kaiming He, Xiangyu Zhang, Shaoqing Ren, and Jian Sun.
\newblock Delving deep into rectifiers: Surpassing human-level performance on
  imagenet classification.
\newblock In {\em Proceedings of the IEEE international conference on computer
  vision}, pages 1026--1034, 2015.

\bibitem{pennington2014glove}
Jeffrey Pennington, Richard Socher, and Christopher Manning.
\newblock Glove: Global vectors for word representation.
\newblock In {\em Proceedings of the 2014 conference on empirical methods in
  natural language processing (EMNLP)}, pages 1532--1543, 2014.

\bibitem{chiu2016named}
Jason~PC Chiu and Eric Nichols.
\newblock Named entity recognition with bidirectional lstm-cnns.
\newblock {\em Transactions of the Association for Computational Linguistics},
  4:357--370, 2016.

\bibitem{lample2016neural}
Guillaume Lample, Miguel Ballesteros, Sandeep Subramanian, Kazuya Kawakami, and
  Chris Dyer.
\newblock Neural architectures for named entity recognition.
\newblock {\em arXiv preprint arXiv:1603.01360}, 2016.

\bibitem{ma2016end}
Xuezhe Ma and Eduard Hovy.
\newblock End-to-end sequence labeling via bi-directional lstm-cnns-crf.
\newblock {\em arXiv preprint arXiv:1603.01354}, 2016.

\bibitem{le2018convolutional}
Hoa~T Le, Christophe Cerisara, and Alexandre Denis.
\newblock Do convolutional networks need to be deep for text classification?
\newblock In {\em Workshops at the Thirty-Second AAAI Conference on Artificial
  Intelligence}, 2018.

\bibitem{zhang2016highway}
Yu~Zhang, Guoguo Chen, Dong Yu, Kaisheng Yaco, Sanjeev Khudanpur, and James
  Glass.
\newblock Highway long short-term memory rnns for distant speech recognition.
\newblock In {\em 2016 IEEE International Conference on Acoustics, Speech and
  Signal Processing (ICASSP)}, pages 5755--5759. IEEE, 2016.

\bibitem{mueller2016siamese}
Jonas Mueller and Aditya Thyagarajan.
\newblock Siamese recurrent architectures for learning sentence similarity.
\newblock In {\em Thirtieth AAAI Conference on Artificial Intelligence}, 2016.

\bibitem{wang2016compare}
Shuohang Wang and Jing Jiang.
\newblock A compare-aggregate model for matching text sequences.
\newblock {\em arXiv preprint arXiv:1611.01747}, 2016.

\bibitem{chen2016enhanced}
Qian Chen, Xiaodan Zhu, Zhenhua Ling, Si~Wei, Hui Jiang, and Diana Inkpen.
\newblock Enhanced lstm for natural language inference.
\newblock {\em arXiv preprint arXiv:1609.06038}, 2016.

\bibitem{wang2017bilateral}
Zhiguo Wang, Wael Hamza, and Radu Florian.
\newblock Bilateral multi-perspective matching for natural language sentences.
\newblock {\em arXiv preprint arXiv:1702.03814}, 2017.

\bibitem{wang2018deep}
Mei Wang and Weihong Deng.
\newblock Deep visual domain adaptation: A survey.
\newblock {\em Neurocomputing}, 312:135--153, 2018.

\end{thebibliography}

\end{document}